\pdfoutput=1
\documentclass[sigconf]{acmart}

\usepackage{booktabs}
\usepackage{subcaption}
\usepackage{multirow}
\usepackage{array}
\usepackage{xcolor}
\usepackage{amsmath}
\usepackage{dsfont}

\newcommand{\udfsection}[1]{\vspace{1mm}\noindent\textbf{#1}.}
\newcommand{\vech}{\textbf{h}}
\newcommand{\vecr}{\textbf{r}}
\newcommand{\vect}{\textbf{t}}
\newcommand{\vecR}{\textbf{R}}

\newcommand{\hlcolor}{\color{black}}

\newcommand{\modify}[1]{{\hlcolor#1}}
\AtBeginDocument{%
  \providecommand\BibTeX{{%
    \normalfont B\kern-0.5em{\scshape i\kern-0.25em b}\kern-0.8em\TeX}}}

\copyrightyear{2022} 
\acmYear{2022} 
\setcopyright{acmcopyright}\acmConference[WWW '22]{Proceedings of the ACM Web Conference 2022}{April 25--29, 2022}{Virtual Event, Lyon, France}
\acmBooktitle{Proceedings of the ACM Web Conference 2022 (WWW '22), April 25--29, 2022, Virtual Event, Lyon, France}
\acmPrice{15.00}
\acmDOI{10.1145/3485447.3511923}
\acmISBN{978-1-4503-9096-5/22/04}

\begin{document}

\title[Rethinking Graph Convolutional Networks in Knowledge Graph Completion]{Rethinking Graph Convolutional Networks in Knowledge\\ Graph Completion}

\author{Zhanqiu Zhang$^1$, Jie Wang$^{1,2,\ast}$, Jieping Ye$^{3,4}$, Feng Wu$^{1}$}
\thanks{$\ast$ Corresponding Author}
\affiliation{%
\institution{$^1$CAS Key Laboratory of Technology in GIPAS\\
University of Science and Technology of China}
 \institution{
$^2$Institute of Artificial Intelligence\\
Hefei Comprehensive National Science Center}
\country{}
 }
 \email{zqzhang@mail.ustc.edu.cn, {jiewangx, fengwu}@ustc.edu.cn}
\affiliation{%
\institution{$^3$University of Michigan}
\institution{$^4$KE Holdings Inc.}
\country{}
}
\email{jpye@umich.edu}

\renewcommand{\shortauthors}{Zhang et al.}

\begin{abstract}
Graph convolutional networks (GCNs)---which are effective in modeling graph structures---have been increasingly popular in knowledge graph completion (KGC). GCN-based KGC models first use GCNs to generate expressive entity representations and then use knowledge graph embedding (KGE) models to capture the interactions among entities and relations. However, many GCN-based KGC models fail to outperform state-of-the-art KGE models though introducing additional computational complexity. This phenomenon motivates us to explore the real effect of GCNs in KGC. Therefore, in this paper, we build upon representative GCN-based KGC models and introduce variants to find which factor of GCNs is critical in KGC. Surprisingly, we observe from experiments that the graph structure modeling in GCNs does not have a significant impact on the performance of KGC models, which is in contrast to the common belief. Instead, the transformations for entity representations are responsible for the performance improvements. Based on the observation, we propose a simple yet effective framework named LTE-KGE, which equips existing KGE models with linearly transformed entity embeddings. Experiments demonstrate that LTE-KGE models lead to similar performance improvements with GCN-based KGC methods, while being more computationally efficient. These results suggest that existing GCNs are unnecessary for KGC, and novel GCN-based KGC models should count on more ablation studies to validate their effectiveness. The code of all the experiments is available on GitHub at \url{https://github.com/MIRALab-USTC/GCN4KGC}.
\end{abstract}



\maketitle

\section{Introduction}
Knowledge graphs (KG) contains quantities of factual triplets---(head entity, relation, tail entity)---which are widely used in many areas, such as e-commerce \cite{alicoco,alicoco2}, finance \cite{finance_kg,kbert}, and social networks \cite{kgat,scholar_kg}. Since knowledge graphs are usually incomplete and manually finding all factual triplets is expensive, how to automatically perform knowledge graph completion (KGC) has attracted great attention. In the past few years, many knowledge graph completion models have been proposed. For example, knowledge graph embeddings (KGE) \cite{transe,distmult,complex,conve,rotate,cone} are popular KGC models due to their simplicity and effectiveness. KGE models embed entities and relations in a knowledge graph as low-dimensional representations (embeddings) and define scoring functions on embedding spaces to measure the plausibility of triplets. In general, valid triplets are supposed to have higher scores than invalid triplets.

\begin{figure*}[ht]
    \centering
    \includegraphics[width=1.7\columnwidth]{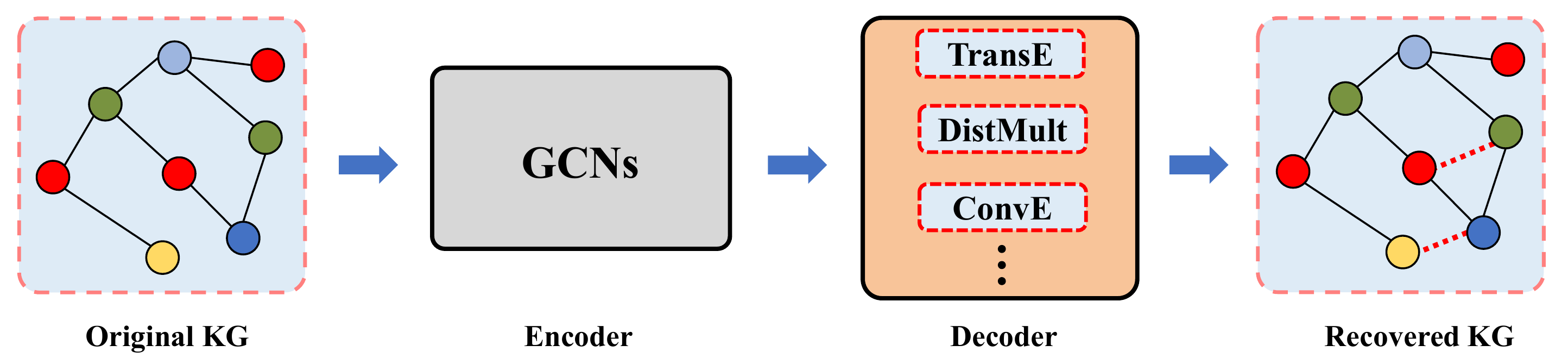}
    \vspace{-2mm}
    \caption{The encoder-decoder framework of GCN-based KGC. GCNs are encoders to generate entity and relation representations based on graph structure. KGE models are decoders to recover the structure of knowledge graphs (KG) according to the entity and relation representations generated by encoders. The predicted unseen links are marked by red dashed line.}
    \label{fig:en-de}
    \vspace{-2mm}
\end{figure*}

Recently, graph convolutional network (GCNs) have been increasingly popular \cite{rgcn,wgcn,vec_kg,kbgat,compgcn} in KGC, as GCNs are effective in modeling graph structures. 
As shown in Figure \ref{fig:en-de}, GCN-based KGC models usually use an encoder-decoder framework, in which GCNs and KGE models serve as encoders and decoders, respectively. Specifically, GCNs generate expressive representations for entities and relations based on their neighbor entities and relations; KGE models (such as TransE \cite{transe}, DistMult \cite{distmult}, and ConvE \cite{conve}) model the interactions among entities and relations using the representations generated by GCNs. Notably, different from classical GCNs that only aggregate information from neighbor nodes \cite{gcn,graphsage}, GCNs in KGC take edges (relations) in knowledge graphs into account. Many works \cite{rgcn,wgcn,compgcn} have demonstrated that adding GCNs before KGE models leads to better performance.

However, though researchers have made persistent efforts to optimize the architectures and training strategies of GCNs, many GCN-based KGC models fail to outperform state-of-the-art KGE models. For example, the state-of-the-art GCN-based KGC model CompGCN \cite{compgcn} achieves mean reciprocal rank (MRR, higher is better) of 0.355 and 0.479 on the FB15k-237 \cite{conve} and WN18RR \cite{wn18rr} datasets respectively, while the state-of-the-art KGE model ComplEx+N3 \cite{n3} attains MRR of 0.366 and 0.489 \cite{dura}. That is to say, GCN-based models do not show great advantages compared with KGE models while introducing additional computational complexity. This phenomenon raises a question worth consideration:  \textit{"What is the real effect of GCNs in the KGC task?"}

In this paper, we try to explore the real effect of GCNs in KGC via extensive experiments. We build upon representative GCN-based KGC models and introduce variants to find which factor of GCNs is critical in KGC. Surprisingly, our experiments demonstrate that randomly breaking adjacency tensors used in GCNs does not have a significant impact on the performance of GCN-based KGC models. By further noticing a bijection between adjacency tensors and graph structures, we conclude that while GCNs improve the performance of KGE models, the graph structure modeling in GCNs does not account for the performance gain. The result overturns the consensus that the main advantage of GCNs is the ability to model graph structures. Besides, we test the effect of the usage of neighbor information and self-loop information, and transformations for relations. Finally, we find that so long as a GCN can distinguish entities with different semantics by its generated entity representations, the transformations for entity representations effectively improve the performance of KGE models. Based on the observations, we propose a simple yet effective framework named LTE-KGE, which equips existing KGE models with linearly transformed entity representations. Experiments demonstrate that LTE-KGE models lead to similar performance improvements to KGE models compared with GCN-based models, while avoiding heavy computational loads in GCN aggregation. Further, we show that though LTE-KGE models do not explicitly aggregate information from graph neighborhoods, they behave like a single GCN layer. These results suggest that existing GCNs are unnecessary for the KGC task, and we should conduct more ablation studies when designing complicated GCN architectures for KGC. 


\section{Preliminaries}
%
\subsection{Knowledge Graphs (KGs)}\label{sec:pre_kg}
Given a set of entities $\mathcal{E}$ and a set of relations $\mathcal{R}$, a knowledge graph $\mathcal{K}=\{(e_i,r_j,e_k)\}\subset \mathcal{E}\times\mathcal{R}\times\mathcal{E}$ is a set of triplets, where $e_i$ and $r_j$ are the $i$-th entity and $j$-th relation, respectively. Usually, we call $e_i$ and $e_k$ the head entity and tail entity, and use $h_i$ and $t_k$ to distinguish them. 
A knowledge graph can be uniquely determined by a third-order binary tensor $\mathcal{X}\in\{0,1\}^{|\mathcal{E}|\times|\mathcal{R}|\times|\mathcal{E}|}$, which is also called the adjacency tensor of $\mathcal{K}$. The $(i,j,k)$ entry $\mathcal{X}_{ijk}=1$ if $(h_i, r_j, t_k)$ is valid or otherwise $\mathcal{X}_{ijk}=0$. 

\subsection{Knowledge Graph Completion (KGC)}\label{sec:pre_kgc}
The goal of KGC is to predict valid but unobserved triplets based on known triplets in $\mathcal{K}$. KGC models define a score function $s:\mathcal{E}\times\mathcal{R}\times\mathcal{E}\rightarrow \mathbb{R}$ to associate a score $s(h_i,r_j,t_k)$ with each potential triplet $(h_i,r_j,t_k)\in\mathcal{E}\times\mathcal{R}\times\mathcal{E}$. The scores measure the plausibility of triplets. For a query $(h_i,r_j,?)$ or $(?,r_j,
t_k)$, KGC models first fill the blank with each entity in the knowledge graphs and then score the resulted triplets. Valid triplets are expected to have higher scores than invalid triplets.

Knowledge graph embedding (KGE) models are popular KGC models, which associate each entity $e_i\in\mathcal{E}$ (either be head or tail entities) and relation $r_j\in\mathcal{R}$ with an embedding $\textbf{e}_i$ and $\vecr_j$ on carefully picked embedding spaces. Then, they directly define a scoring function to model the interactions among entities and relations. We review three representative KGE models as follows.

\subsubsection{TransE \cite{transe}}
TransE uses the Minkowski distances to define scoring functions on the real space. Specifically, given the embeddings $\vech,\vecr,\vect$, the scoring function is 
\begin{align*}
    f(h,r,t)=-\|\vech+\vecr-\vect\|_{1/2},
\end{align*}
where $\|\cdot\|_{1/2}$ means that the distance can be $L_1$ or $L_2$ distance.

\subsubsection{DistMult \cite{distmult}}
DistMult uses the inner product to define scoring functions on the real space. Specifically, given the embeddings $\vech,\vecr,\vect$, the scoring function is 
\begin{align*}
    f(h,r,t)=\vech^\top\textbf{R} \vect,
\end{align*}
where $\textbf{R}$ is a diagonal matrix whose diagonal elements are $\vecr$.

\begin{table*}[ht]
    \centering
    \caption{The embedding update processes of three popular GCNs in KGC. For a given entity $e$, the set $\mathcal{N}_{in}(e)$ consists of all head entity and relation pairs $(h,r)$ such that $(h,r,e)$ is valid. The set $\mathcal{N}_{out}(e)$ consists of all relation and tail entity pairs $(r,t)$ such that $(e,r,t)$ is valid. $\textbf{W}$ are trainable linear transformations. $\phi_{in}/\phi_{out}$ are direction-dependent composition operators.}
    \label{tab:emb_update}
    \vspace{-1.5mm}
    \begin{tabular}{cccc}
    \toprule
    & \textbf{Entity Aggregation} & \textbf{Entity Update} & \textbf{Relation}\\
    \midrule
    RGCN \cite{rgcn}  & $\textbf{m}_e^{l+1}=\sum_{(h,r)\in\mathcal{N}_{in}(e)} \textbf{W}_r^l \vech^l + \sum_{(r,t)\in\mathcal{N}_{out}(e)} \textbf{W}_r^l \vect^l+\textbf{W}_0^l \textbf{e}^l$ &$\textbf{e}^{l+1}=\sigma(\textbf{m}_e^{l+1})$ & $\vecr^{l+1}=\vecr^l$\\
    WGCN \cite{wgcn}     & $\textbf{m}_e^{l+1}=\sum_{(h,r)\in\mathcal{N}_{in}(e)} \textbf{W}^l (\alpha_r^l \vech^l) + \sum_{(r,t)\in\mathcal{N}_{out}(e)} \textbf{W}^l (\alpha_r^l \textbf{t}^l)+\textbf{W}_0^l \textbf{e}^l$ & $\textbf{e}^{l+1}=\sigma(\textbf{m}_e^{l+1})$ &$\vecr^{l+1}=\vecr^l$\\
    CompGCN \cite{compgcn} & $\textbf{m}_e^{l+1}=\sum_{(h,r)\in\mathcal{N}_{in}(e)} \textbf{W}_r^l \phi_{in}(\vech^l,\vecr^l)+ \sum_{(r,t)\in\mathcal{N}_{out}(e)} \textbf{W}_r^l \phi_{out}(\vect^l,\vecr^l)+\textbf{W}_0^l\textbf{e}^l$ & $\textbf{e}^{l+1}=\sigma(\textbf{m}_e^{l+1})$ &$\vecr^{l+1}=\textbf{W}_{rel}^l \vecr^l$ \\
    \bottomrule
    \end{tabular}
\end{table*}

\subsubsection{ConvE \cite{conve}}
ConvE uses convolutional neural networks to define scoring functions. The scoring function is 
\begin{align*}
    f(h,r,t)=\sigma(\text{vec}(\sigma([\bar{\vecr}, \bar{\vech}]*\omega))W)^\top \vect,
\end{align*}
where $\sigma$ denotes an activation function, $*$ denotes 2D convolution, $\omega$ denotes a filter in convolutional layers, $\bar{\cdot}$ denotes 2D shaping for real vectors, and $W$ is a trainable weight matrix.

\subsection{Graph Convolutional Network Based KGC}\label{sec:pre_gnnkgc}
Many popular GCNs follow an iterative aggregate and update scheme. In each iteration, node representations are updated using information aggregated from nodes' neighborhoods (both nodes and edges). When applied to KGs, GCNs are modified to model the interactions among entities and relations.
As shown in Figure \ref{fig:en-de}, GCN-based KGC models usually use an encoder-decoder framework \cite{rgcn}, where GCNs perform as the encoder and KGE models (e.g., TransE, DistMult, and ConvE) perform as the decoder. First, the encoder generates representations for entities and relations in a KG. Compared with directly using trainable vectors in embedding spaces, representations generated by GCNs are expected to capture more structure information around entities and relations. Then, the decoder uses the generated representations to predict values in the adjacency tensors. Since there exists a bijection between adjacency tensors and graph structures, the prediction can be seen as a recovery of the original graph structures. When recovering graph structure, the decoder can predict links missing in the original graph, i.e., complete the knowledge graph. Note that KGE models can also be interpreted under this encoder-decoder framework, where the encoder is an identity function from the embedding space and the decoder is the KGE model itself.

\section{Experimental Setup}\label{sec:exp_setup}
In this section, we introduce our experimental setups, including models, loss functions, datasets, and training/evaluation protocols.

\subsection{Models}\label{sec:exp_models}
To investigate the effect of GCNs (encoders) in KGC, we study three representative models, including \textbf{Relational-GCN (RGCN)} \cite{rgcn}, \textbf{Weighted-GCN (WGCN)} \cite{wgcn}, and \textbf{CompGCN} \cite{compgcn}, all of which are specially designed for multi-relational graphs. Specifically, RGCN extends vanilla GCN with relation-specific linear transformations, without considering relation representations. WGCN treats the relation as learnable weights of edges and applies vanilla GCN on the simple weighted graph. CompGCN jointly embeds both entities and relations in knowledge graphs and uses composition operators to model the interactions among entities and relations.
The embedding update processes of these three models are summarized in Table \ref{tab:emb_update}. We can see that the embedding update processes have three main parts: 1) aggregating information from graph neighborhoods, 2) transformations for aggregated entity representations, and 3) transformations for relation representations.

As to KGE models (decoders), we use three popular models: \textbf{TransE}, \textbf{DistMult}, and \textbf{ConvE} (T, D, C for short).  We choose $L_1$ norm for TransE to achieve its best performance.

\subsection{Loss Functions}\label{sec:exp_loss}
Many existing state-of-the-art GCN-based KGC models \cite{rgcn,wgcn,compgcn} use the following binary cross entropy (BCE) loss 
\begin{align}\label{eqn:loss}
    L&=\frac{1}{|S|}\sum_{(h,r,\cdot)\in S}\left(\frac{1}{|\mathcal{E}|}\sum_{t\in \mathcal{E}}y_{(h,r,t)}\cdot\log f(h,r,t)\right.\nonumber\\
    &\left.+(1-y_{(h,r,t)})\cdot\log (1-f(h,r,t))\vphantom{\sum_{t\in \mathcal{E}}}\right),
\end{align}
where $S$ is the set consisting of all known valid triplets, $y_{(h,r,t)}$ is the label of the triplet $(h,r,t)$. Label smoothing is used to prevent models from overfitting. The labels $y_{(h,r,t)}$ are smoothed to be numbers between $[0,1]$. Using this loss, we regard all the triplets not seen in the training set as negative samples.

\subsection{Datasets}\label{sec:exp_datasets}
We use two popular knowledge graph completion datasets FB15k-237 (FB237) and WN18RR, which are subsets of FB15k \cite{transe} and WN18 \cite{transe}, respectively.
Since FB15k and WN18 suffer from the test set leakage problem \cite{conve,wn18rr}, we do not include the results on FB15k and WN18 in the experiments. The statistics of these two datasets are summarized in Appendix \ref{app:data}.

\subsection{Training and Evaluation Protocols}\label{sec:exp_protocol}
\udfsection{Training}
We re-implement RGCN, WGCN, and CompGCN with DGL \cite{dgl}.
As we aim to isolate the effects of GCNs instead of benchmarking the performance of GCN-based KGC models, we follow the guidelines provided by the original GCN papers \cite{rgcn,wgcn,compgcn} or official implementations for model selection and do not spend time in tuning hyper-parameters. 
Since there is no suggested hyper-parameters for RGCN on WN18RR in the original papers or implementations, we omit the corresponding results. Nonetheless, it does not affect our overall conclusions.
\modify{
Notably, further tuning hyper-parameters does not affect our conclusions. First, for original GCN models, since we use their suggested best hyper-parameters, the performance of re-implemented models is comparable to the reported performance in \citet{compgcn} (see Section \ref{sec:repro}). 
Second, for model variants, so long as they achieve competitive performance with the original models, we can make our conclusion. The performance of current implementations has already met our demands. Further tuning hyper-parameters for these variants may improve their performance, while it makes no difference to the conclusion. Besides, the strategy that does not tune hyper-parameters for model variants is widely used in works revisiting existing models \cite{rethink1,rethink2,rethink3}. We follow their settings in our experiments.
}

\begin{table}[ht]
    \centering
    \caption{Comparison of MRR between the state-of-the-art KGE and GCN-based KGC models. *, \textdagger, and $\diamond$ indicate that the results are taken from \cite{old_dog}, \cite{dura}, and \cite{compgcn}, respectively.}
    \label{tab:comp_gnn_kge}
    \vspace{-1.5mm}
    \begin{tabular}{ccc}
        \toprule
       & FB237  & WN18RR \\
        \midrule
        TransE*  & .288 & .202\\
        DistMult*  & .347 & .447\\
        ConvE*  & .344 & .447  \\
        ComplEx*  & .351 & .477\\
        ComplEx-N3\textdagger  & \textbf{.366} & \textbf{.489}\\
        \midrule
        RGCN+ConvE$\diamond$  & .342 & -\\
        WGCN+ConvE$\diamond$ & .344 & - \\
        CompGCN+ConvE$\diamond$ & .355 & .479 \\
        \bottomrule
    \end{tabular}
    \vspace{-2.5mm}
\end{table}

\udfsection{Evaluation}
For each triplet $(h,r,t)$ in test sets, we replace either $h$ or $t$ with candidate entities to create candidate triplets. We then rank the candidate triplets in descending order by their scores \cite{transe}. We use the ``Filtered'' setting \cite{transe} that does not take existing valid triplets into accounts at ranking. We use Mean Rank (MR),  Mean Reciprocal Rank (MRR), and Hits at N (H@N) as the evaluation metrics. Lower MR and higher MRR/H@N indicate better performance. 

\section{Real effect of GCNs in KGC}\label{sec:analysis}
GCNs can generate expressive node representations, which is supposed to benefit downstream tasks. However, as shown in Table \ref{tab:comp_gnn_kge}, GCN-based KGC models do not show a great advantage over state-of-the-art KGE models though introducing additional computational complexity. For example, the state-of-the-art GCN model CompGCN \cite{compgcn} achieves MRR of 0.355 and 0.479 on FB237 \cite{conve} and WN18RR \cite{wn18rr} respectively, while the KGE model ComplEx+N3 \cite{n3} attains MRR of 0.366 and 0.489 \cite{dura}. The results raise  a natural question: "\textit{What is the real effect of GCNs in the KGC task?}"
To answer the question, we decompose it into the following two sub-questions.
\begin{enumerate}
    \item Do GCNs really bring performance gain?
    \item Which factor of GCNs is critical in KGC?
\end{enumerate}

\subsection{Do GCNs Really Bring Performance Gain?}\label{sec:repro}
Some GCN-based KGC works show impressive performance gain \cite{rgcn,wgcn,compgcn}, while the baseline results are taken from the original papers of KGE models. As advanced training strategies can significantly boost the performance of KGE models \cite{old_dog} and baseline KGE models do not use these strategies, the benefit of GCNs may be overestimated. Thus, we conduct experiments to validate whether GCNs really bring performance gain over KGE models.

\begin{table}[ht]
    \caption{Comparison of MRR between models without (w/o) and with GCNs.  "O" indicates that the results are taken from their original papers, while "R" indicates that we reproduce the results using DGL. T: TransE. D: DistMult. C: ConvE.}\label{tab:comp_wognn}
    \centering
    \vspace{-1mm}
    \begin{tabular}{c*3{c}*3{>{\hlcolor}c}}
    \toprule
     &\multicolumn{3}{c}{FB237} &\multicolumn{3}{c}{\hlcolor WN18RR}\\
    \cmidrule(lr){2-4}
    \cmidrule(lr){5-7}
    & T & D  & C & T & D  & C\\
    \midrule
    w/o GCN (O)  & .294 &  .241  &.325 &- &- &-\\
    RGCN (O) & .281& .324  &.342&- &- &-  \\
    WGCN (O) & .267& .324  &.344 &- &- &- \\
    CompGCN (O)  & .337& .338  & .353 &- &- &.479\\
    \midrule
    w/o GCN (R)  & .332&  .279  &.319&.205 &.410 &.462\\
    RGCN (R)  & .324 & .332 &.337&-&-&- \\
    WGCN (R)  &.272& .329  &.340 &.222 &.422 &.462\\
    CompGCN (R)  & .335& .342  & .353 &.206 &.430 &.469\\
    \bottomrule
    \end{tabular}
\end{table}

Table \ref{tab:comp_wognn} show that in most cases, GCNs, especially the state-of-the-art CompGCN, significantly improve the performance of KGE models. Notably, on FB237, WGCN+TransE performs worse than TransE in both the original and reproduced cases. A similar phenomenon can also be observed for WGCN+DistMult/ConvE on WN18RR. It demonstrates that not all GCNs can improve the performance of all KGE models. Moreover, the reproduced performance of DistMult and TransE is significantly better than their reported performance, showing the influence of different training strategies. 

In summary, the answer to the first question is "Yes". GCNs do bring performance gain over KGE models for the KGC task.

\begin{table}[ht]
\caption{MRR for GCNs with random adjacency tensors (RAT) and without neighbor information (WNI). }
    \label{tab:comp_rat_wni}
    \centering
    \vspace{-1.5mm}
    \begin{tabular}{c*3{c}*3{>{\hlcolor}c}}
    \toprule
    &\multicolumn{3}{c}{FB237} &\multicolumn{3}{c}{\hlcolor WN18RR} \\
    \cmidrule(lr){2-4}
    \cmidrule(lr){5-7}
    & T & D  & C & T & D  & C\\
    \midrule
    RGCN & .324 & .332 &.337&-&-&- \\
    RGCN + RAT &.322& .331  &.333 &-&-&- \\
    RGCN + WNI  &.324&.332  &.335 &-&-&-\\
    \midrule
    WGCN  &.272& .329  &.340 &.222 &.422 &.462\\
    WGCN + RAT &.325& .330  &.354  &.213 &.426 &.470\\
    WGCN + WNI  &.337&.334  &.355 &.235 &.425 &.473\\
    \midrule
    CompGCN & .335& .342  & .353 &.206 &.430 &.469\\
    CompGCN + RAT  & .336&.336  & .351 &.208 &.420 &.467\\
    CompGCN + WNI  & .339&.335 & .352 &.214 &.435 &.465\\
    \bottomrule
    \end{tabular}
    \vspace{-1.5mm}
\end{table}

\subsection{Which Factor of GCNs is Critical in KGC?}
There is a consensus that the performance improvements of GCNs mainly comes from their ability to model graph structures. However, whether graph structure modeling of GCNs is critical in KGC is less explored, and which factor of GCNs is critical in KGC is unclear. Thus, we conduct extensive experiments to test the effect of graph structures, neighbor information, self-loop information, and linear transformations for relations.

\subsubsection{Graph Structures}
GCNs are known to be effective in modeling graph structures. Therefore, if we break graph structures, the performance of GCN-based KGC models is expected to decrease significantly. Since graph structure information of a knowledge graph is represented by its adjacency tensors (please refer to Section \ref{sec:pre_kgc} for a detailed definition), we conduct experiments with randomly broken adjacency tensors to explore the effect of the graph structures. Specifically, when building the adjacency tensor for message passing, given a valid triplet $(h_i,r_j,t_k)$, we replace the tail entity $t_k$ with a random entity in the knowledge graph. Note that we only use the random adjacency tensors in message passing, while the training/validation/test triplets remain unchanged.

\begin{table}[ht]
\caption{MRR results for GCNs without self-loop information (WSI) and random adjacency tensors (RAT). 
}\label{tab:comp_wsi}
    \centering
    \vspace{-1.5mm}
    \resizebox{\columnwidth}{!}{
    \begin{tabular}{c*3{c}*3{>{\hlcolor}c}}
    \toprule
    &\multicolumn{3}{c}{FB237} &\multicolumn{3}{c}{\hlcolor WN18RR}\\
    \cmidrule(lr){2-4}
    \cmidrule(lr){5-7}
    & T & D  & C & T & D  & C\\
    \midrule
    RGCN & .324 & .332 &.337&-&-&- \\
    RGCN + WSI &.323 &.337  &.339 &-&-&- \\
    RGCN + WSI+RAT &.320&.317  &.318  &-&-&-\\
    \midrule
    WGCN  &.272& .329  &.340 &.222 &.422 &.462\\
    WGCN + WSI   &.263 &.330&.341 &.181 &.408 &.426\\
    WGCN + WSI + RAT   &.322 &.319&.340 &.168 &.353 &.408\\
    \midrule
    CompGCN & .335& .342  & .353 &.206 &.430 &.469\\
    CompGCN + WSI  & .320  &.338& .352 &.181 &.408 &.426\\
    CompGCN + WSI + RAT  & .317 &.297 & .342 &.168 &.353 &.408\\
    \bottomrule
    \end{tabular}
    }
    \vspace{-1mm}
\end{table}

Table \ref{tab:comp_rat_wni} shows the results for GCNs with random adjacency tensors (X+RAT). Surprisingly, randomly breaking the adjacency tensors, i.e., the graph structures, does not affect the overall performance of GCN-based KGC models \modify{on both datasets}. The models with random adjacency tensors attain comparative performance to their normal adjacency tensor counterparts. For WGCN+TransE, random graph structures even improve the performance on FB237. The results demonstrate that, although GCN encoders can improve the performance of KGE models, \textbf{graph structure modeling in GCNs is not critical for the performance improvements}.

\subsubsection{Neighbor Information}
To further explore the relationship between graph structure modeling in GCNs and the performance improvements, we conduct experiments that do not use neighbor information in the aggregation process. That is to say, the graph used in GCNs has no edges (relations) between nodes (entities), and the new representation of an entity is generated only based on the entity's previous representation.

Table \ref{tab:comp_rat_wni} shows that \modify{on both datasets}, the models without using neighbor information (X+WNI) perform competitively with the original models. It demonstrates that \textbf{the performance gain does not come from the neighborhood aggregation}.

\subsubsection{Self-loop Information}
To determine whether self-loop information is necessary for the performance gain, we conduct experiments without self-loop information. That is to say, the representation of an entity is generated only based on the representations of its neighborhood entities and relations. 

\udfsection{Results on FB237}
Table \ref{tab:comp_wsi} shows the results without self-loop information  (X+WSI). Surprisingly, leaving out the self-loop information does not have a significant impact on the performance of most models as well. 
In most cases, \textbf{only aggregating the neighbor information achieves comparative results to full GCN-based KGC models}.
Further, we randomly break the adjacency tensors while leaving out self-loop information. Since we only use neighbor information, we expect the random adjacency tensors to reduce the performance significantly. However, as shown in Table \ref{tab:comp_wsi} (X+WSI+RAT), the performance is only slightly affected for most of the decoders. That is, \textbf{only aggregating randomly generated neighbor information achieve comparative results to full GCN-based KGC models.}

\begin{figure}[ht]
    \centering
    \begin{subfigure}{0.5\columnwidth}
      \includegraphics[width=122pt]{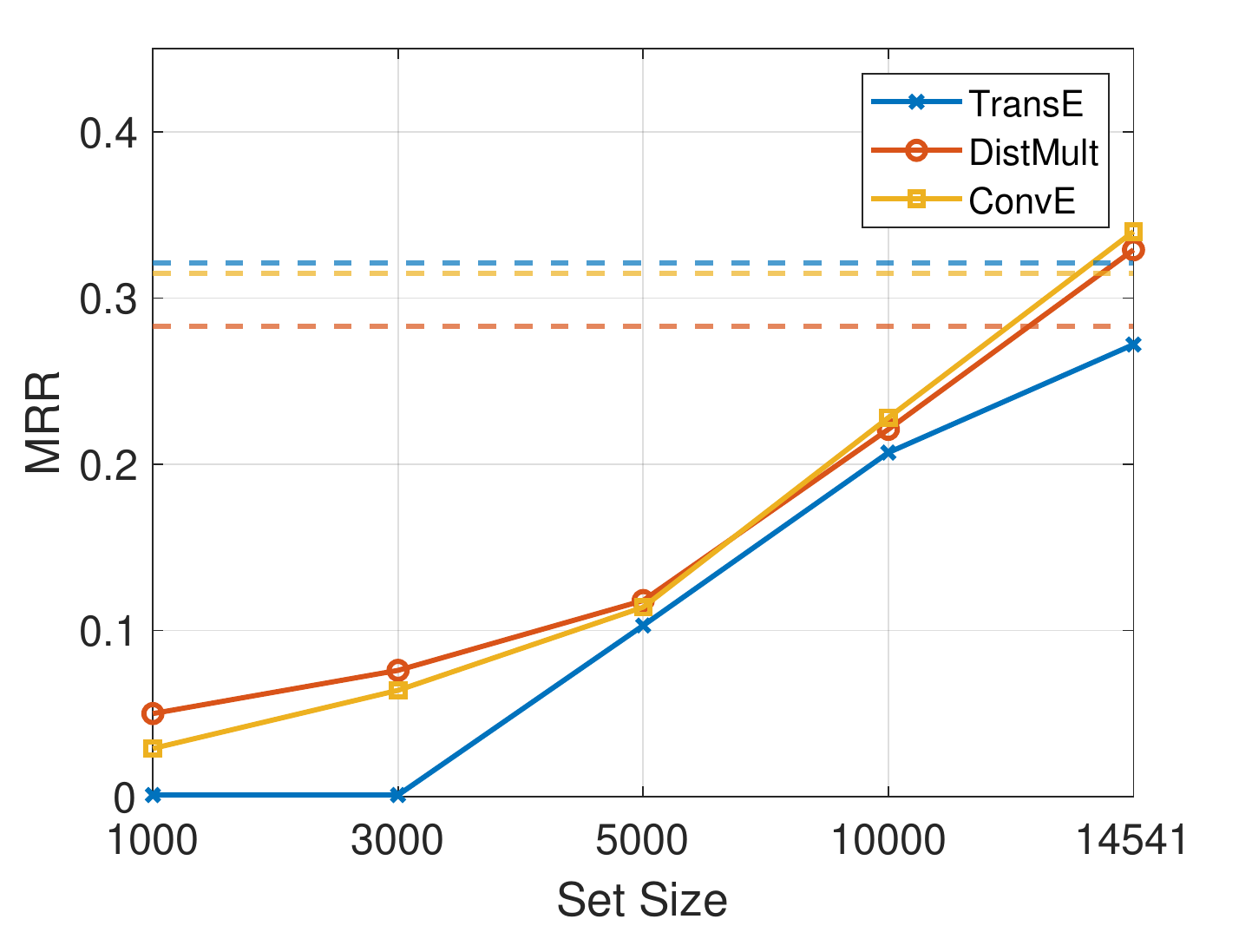}
      \caption{WGCN+X}\label{subfig:sample_size_wgcn}
    \end{subfigure}\hfil
    \begin{subfigure}{0.5\columnwidth}
      \includegraphics[width=122pt]{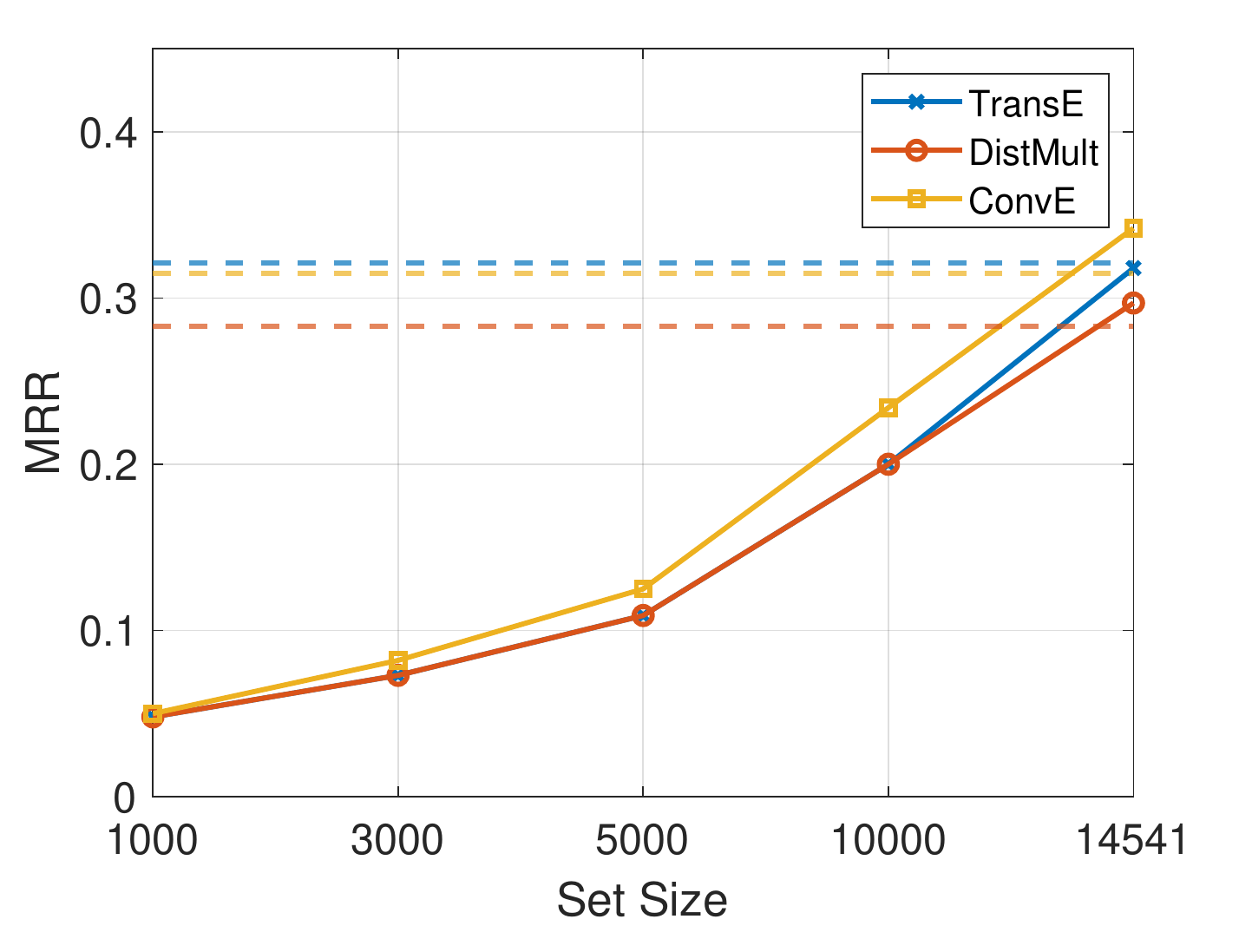}
      \caption{CompGCN+X}\label{subfig:sample_size_compgcn}
    \end{subfigure}\hfil
    \vspace{-1.5mm}
    \caption{MRR results for no self-loop information and random sampled neighbor entities on FB237, where the dash lines indicate performance without GCNs.}
    \label{fig:sample_size}
    \vspace{-1mm}
\end{figure}

\begin{figure}[ht]
    \centering
    \begin{subfigure}{0.5\columnwidth}
      \includegraphics[width=122pt]{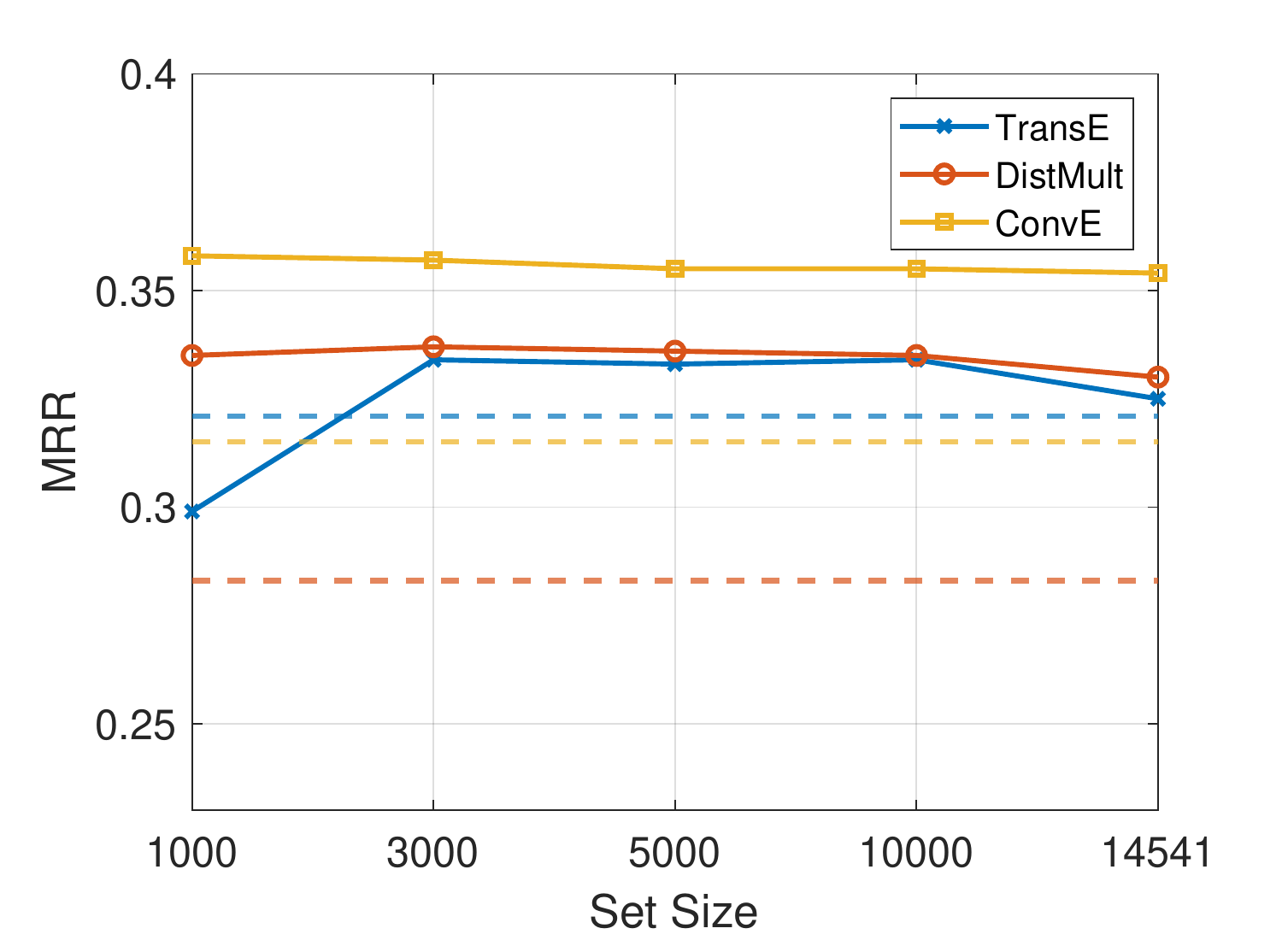}
      \caption{WGCN+X}\label{subfig:sample_size_wloop_wgcn}
    \end{subfigure}\hfil
    \begin{subfigure}{0.5\columnwidth}
      \includegraphics[width=122pt]{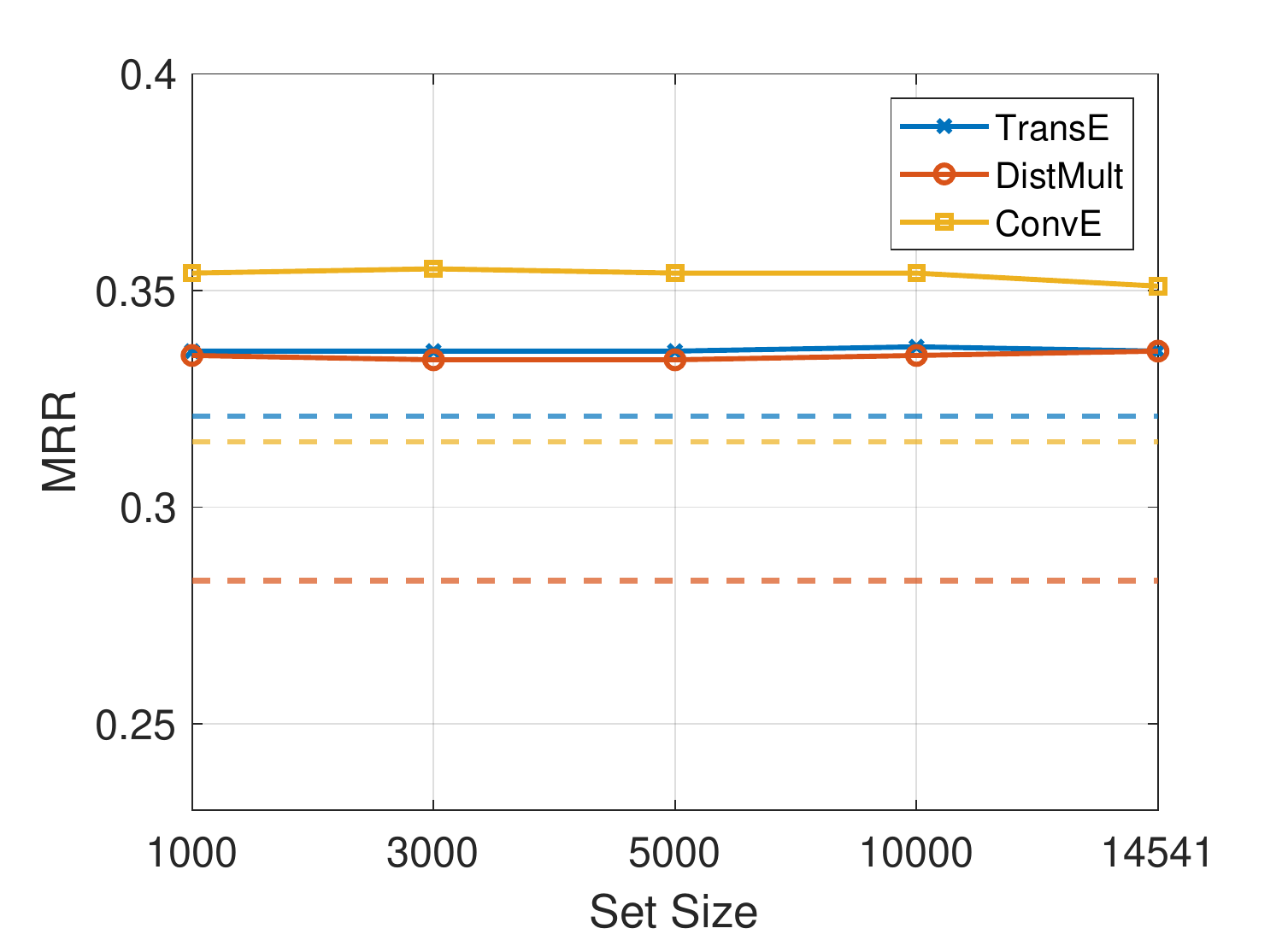}
      \caption{CompGCN+X}\label{subfig:sample_size_wloop_compgcn}
    \end{subfigure}\hfil
    \vspace{-1.5mm}
    \caption{MRR for random sampled neighbors on FB237, where the dash lines indicate performance without GCNs.}\label{fig:sample_size_wloop}
    \vspace{-1.5mm}
\end{figure}

Until now, we have known that the following operations do not have a significant effect on the performance of GCN-based KGC models on FB237: 1) only using self-loop information; 2) only using neighbor information; 3) only using randomly generated neighbor information. The three cases have one common property: \textit{they can distinguish entities having different semantics with high confidence}. Specifically, if we only use self-loop information, the representation of each entity is independent and thus can be distinguished. If we only use neighbor information, two entities will have similar representations only if they have similar neighbor representations, which is consistent with an assumption of KGC: entities with similar neighbors have similar semantics. Therefore, we can distinguish entities with different semantics.  When randomly sampling neighbor entities from all the entities, we assign different neighbors for different entities with a high possibility, and thus we can distinguish different entities by the aggregated entity representations.

To find out whether the above property is necessary for the performance gain, we conduct experiments in which we do not use self-loop information and meanwhile randomly sample neighbors in a given entity set. The number of neighbor entities of each entity remains the same as their original values. If the set size is small, different entities are more likely to have similar neighborhoods and thus have similar representations, which makes them hard to be distinguished. Hence, we expect the performance to decrease when the set size becomes smaller. Figure \ref{fig:sample_size} shows that, as the sample set sizes decrease, the MRR results decrease rapidly and become worse than models without GCNs, which meets our expectations.
We also conduct experiments using self-loop information and randomly sampled neighbors. Figure \ref{fig:sample_size_wloop} shows that the performance is relatively stable when the set sizes vary. It is expectable since the self-loop information itself can uniquely determine an entity, no matter what the neighbor information is.

\modify{
\udfsection{Results on WN18RR}
Compared with the results on FB15k-237, Table \ref{tab:comp_wsi} shows that  only using neighbor information (WSI/WSI+RAT) leads to lower performance than full GCN-based models on WN18RR. It is reasonable since the average numbers of neighbor entities in WN18RR and FB15k-237 are 2.1 and 18.7, respectively. It is more challenging to distinguish entities with different semantics based on fewer neighbor entities. Therefore, the results are consistent with the analyses for FB237.
}

\vspace{1mm}
In summary, \textbf{GCNs improve the performance of KGE models if the aggregation process can well distinguish entities with different semantics}.

\begin{table}[ht]
    \caption{Comparison between CompGCN with and without (w/o) linear transformations for relations (LTR).}\label{tab:comp_rellt}
    \centering
    \vspace{-1.5mm}
    \begin{tabular}{ccccc}
    \toprule
    & & TransE&DistMult  & ConvE \\
    \midrule
    \multirow{ 2}{*}{FB237} &with LTR  & 0.335&0.342 & 0.353 \\
    &w/o LTR  & 0.336& 0.343 & 0.352 \\
    \midrule
    \multirow{ 2}{*}{WN18RR} &with LTR& 0.206 &  0.430  &0.469\\
     &w/o LTR & 0.190&  0.433  &0.468\\
    \bottomrule
    \end{tabular}
    \vspace{-2.5mm}
\end{table}
\subsubsection{Linear Transformations for Relations.} 
Different from RGCN and WGCN, CompGCN applies linear transformations for relation embeddings. We conduct ablation experiments to explore the effect of the transformations. Table \ref{tab:comp_rellt} shows that removing the linear transformations for relations does not significantly affect the performance except for CompGCN+TransE on WN18RR. Note that TransE is sensitive to hyper-parameters, and we do not find the best hyper-parameters by grid search. The performance of CompGCN+TransE may be underestimated. Thus, we can conclude that linear transformations for relations are not critical for GCN-based KGC models.
Recall that the embedding update process of a GCN-based KGC model has three main parts (Section \ref{sec:exp_models}). We have shown that two of them---the aggregation based on graph structures and transformations for relations---are not critical for GCN-based KGC models. Thus, \textbf{the transformations for aggregated entity representations are critical for the performance improvements}.

Based on the above results, we conclude that: so long as a GCN can well distinguish entities with different semantics by its generated entity representations, the transformations for entity representations can effectively improve the performance of KGE models.

\section{A simple yet effective framework}
Based on the observations in the last section, we propose a simple yet effective KGC framework, namely LTE-KGE, which uses linearly transformed entity representations to achieve comparative performance to GCN-based KGC models.
\modify{
It is worth noting that we do not aim to propose a new state-of-the-art KGC model. Instead, we aim to demonstrate that simpler models can achieve similar performance to state-of-the-art GCN-based models, concluding that existing complicated GCNs may be unnecessary for KGC. 
}

\subsection{Model Details for LTE-KGE}
Suppose that $f(\vech,\vecr,\vect)$ is a scoring function for the triplet $(h,r,t)$, which corresponds to an existing KGE model. 
Motivated by the previous experimental results, we propose a framework named LTE-KGE, which applies linear transformations to entity representations. Formally, the proposed framework is
\begin{align}\label{eqn:LTE-KGE}
    f(g_h(W_h\vech), \vecr,g_t(W_t\vect))),
\end{align}
where $W_h$ and $W_t$ are linear transformations with trainable weights. We also introduce optional operations $g_h$ and $g_t$, which can be compositions of functions from the function set \{the identity function, non-linear activation functions, batch normalization, dropout\}. These operations correspond to possible non-linear transformations in GCN-based models. Note that: a) the linear transformations $W_h$ and $W_t$ can share the same parameters depending on the experimental results; b) $g_h$ and $g_t$ can be compositions of different functions depending on the experimental results; c) as each entity has its own representations, LTE-KGE can distinguish entities with different semantics; d) when $W_h,W_t$ are identity matrices and $g_h,g_t$ are identity functions, LTE-KGE models recover KGE models.

Compared with GCNs only using self-loop information, LTE-KGE is more flexible to incorporate different transformations for head and tail entities. Compared with TransR \cite{transr} that applies relation-specific projections to entity embeddings, LTE-KGE is more memory-efficient and flexible to incorporate non-linear operations.

\subsection{Experimental Results}
We conduct experiments for DistMult, TransE, and ConvE. Specifically, $W_h$ and $W_t$ are the same, while $g_h$ and $g_t$ are a composition of batch normalization and dropout for DistMult/ConvE, and an identity function for TransE. 

\begin{table*}[ht]
    \caption{Comparison between LTE-KGE models and CompGCN \cite{compgcn}. The symbol \textdagger~ indicates that the results are reproduced with DGL. The reproduced performance of CompGCN is similar to the reported performance in their original paper.}
    \label{tab:simple_results}
    \centering
    \vspace{-1mm}
  \begin{tabular}{l c c c c c   c c c c c }
      \toprule
        &  \multicolumn{5}{c}{\textbf{FB237}}&\multicolumn{5}{c}{\textbf{WN18RR}} \\
      \cmidrule(lr){2-6}
      \cmidrule(lr){7-11}
        & MRR &MR  & H@1  & H@3  & H@10 & MRR &MR  & H@1  & H@3  & H@10\\
        \midrule
        RotatE   & .338 &177 & .241 & .375 & .533  & .476 &3340 & .428 & .492 & .571\\
        TuckER    & .358 &- & .266 & .394 & .544 & .470 &- & .433 & .482 & .526\\
        \midrule
        TransE\textdagger &.332 &182 &.240 &.368 &.516 &.205 &3431 &.022 &.347 &.519 \\
        DistMult\textdagger  &.279 &392 &.202 &.306 & .433 &.410 &7970 &.389 &.420 &.450\\
        ConvE\textdagger & .319 & 276 &.232 &.351 &.492 &.462 &4888 &.431 &.476 &.525 \\
        \midrule
        CompGCN + TransE\textdagger  &.335 &205 &.247 &.369 &.511 &.206 &3182 &.064 &.281 &.502 \\
        CompGCN + DistMult\textdagger &.342 &200 &.252 &.372 & .520 &.430 &4559 &.395 &.439 &.513 \\
        CompGCN + ConvE\textdagger & .353 & 221 &.261 &.388 &.536 &.469 &3065 &.433 &.480 &.543 \\
        \midrule
        LTE-TransE  &.334 &182 &.241 &.370 &.519 &.211 &3290 &.022 &.362 &.521\\
        LTE-DistMult&.335 &238 &.246 &.367 & .517  &.437 &4485 &.403 &.447 &.517 \\
        LTE-ConvE & .355 & 249 &.264 &.389 &.535  &.472 &3434 &.437 &.485 &.544\\
      \bottomrule
  \end{tabular}
\end{table*}

Table \ref{tab:simple_results} shows the comparison results. 
\modify{
Note that the performance of our re-implemented CompGCN is comparable to the reported performance in \cite{compgcn}. Therefore, the comparison between LTE-KGE and GCNs is fair.
}
Overall, LTE-KGE significantly improves the performance of DistMult and ConvE. Although LTE-KGE does not explicitly model local graph structures like GCNs, it performs comparably to the GCN-based KGC models and sometimes even performs better. 
\modify{
We also use RotatE \cite{rotate} and TuckER \cite{tucker} as baselines. The results show that GCN-based models do not consistently show great advantages over these KGE models. One may argue that we can also build LTE-KGE over RotatE/TuckER to achieve better performance. However, as noted before, we propose LTE-KGE to challenge the use of GCNs instead of achieving state-of-the-art performance.
As popular GCN-based models do not use RotatE/TuckER as decoders, we do not build LTE-KGE on top of them either.
}

\begin{figure}[ht]
    \centering
    \begin{subfigure}{0.5\columnwidth}
      \includegraphics[width=110pt]{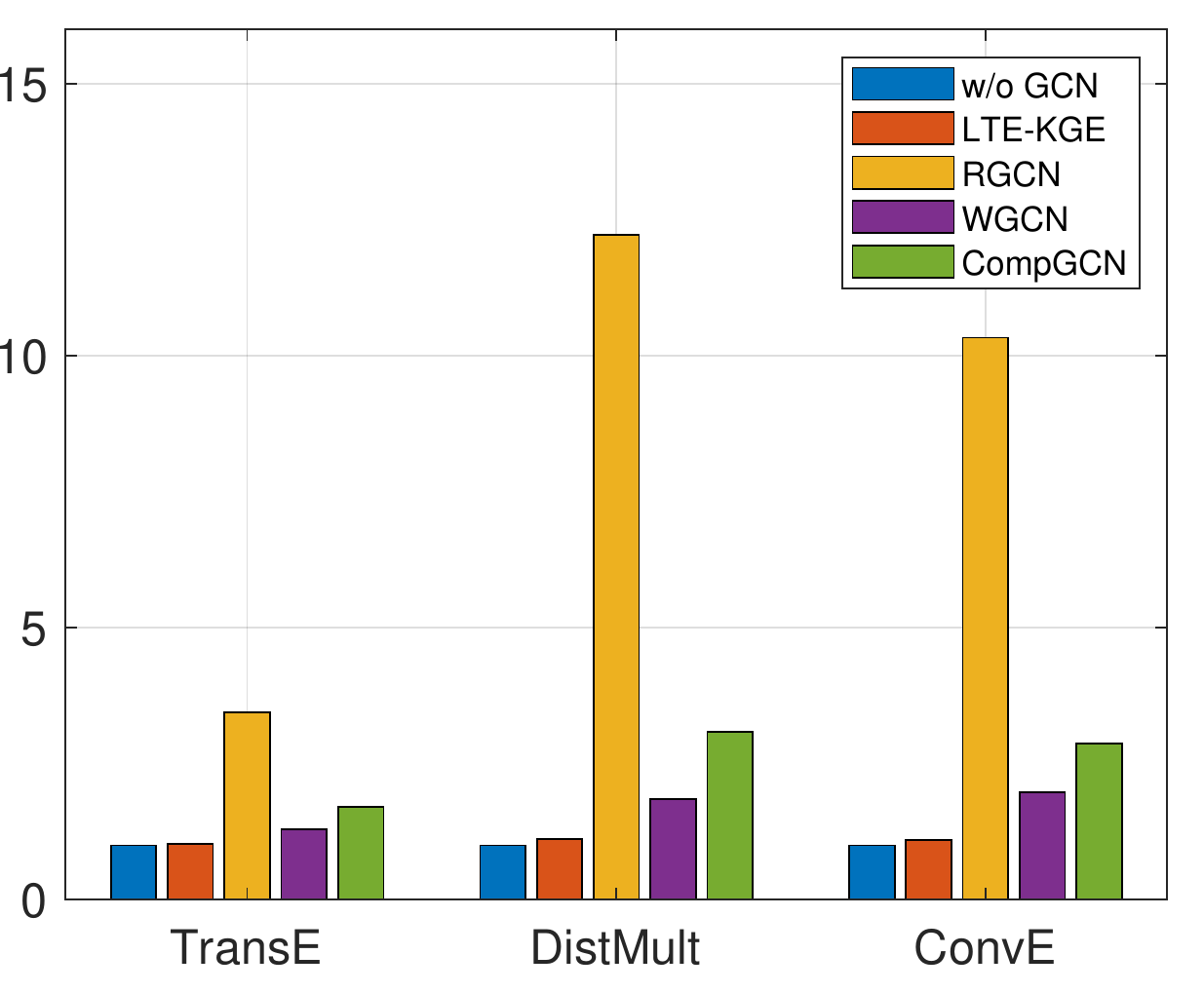}
      \caption{Training Time}\label{subfig:training_time}
    \end{subfigure}\hfil
    \begin{subfigure}{0.5\columnwidth}
      \includegraphics[width=110pt]{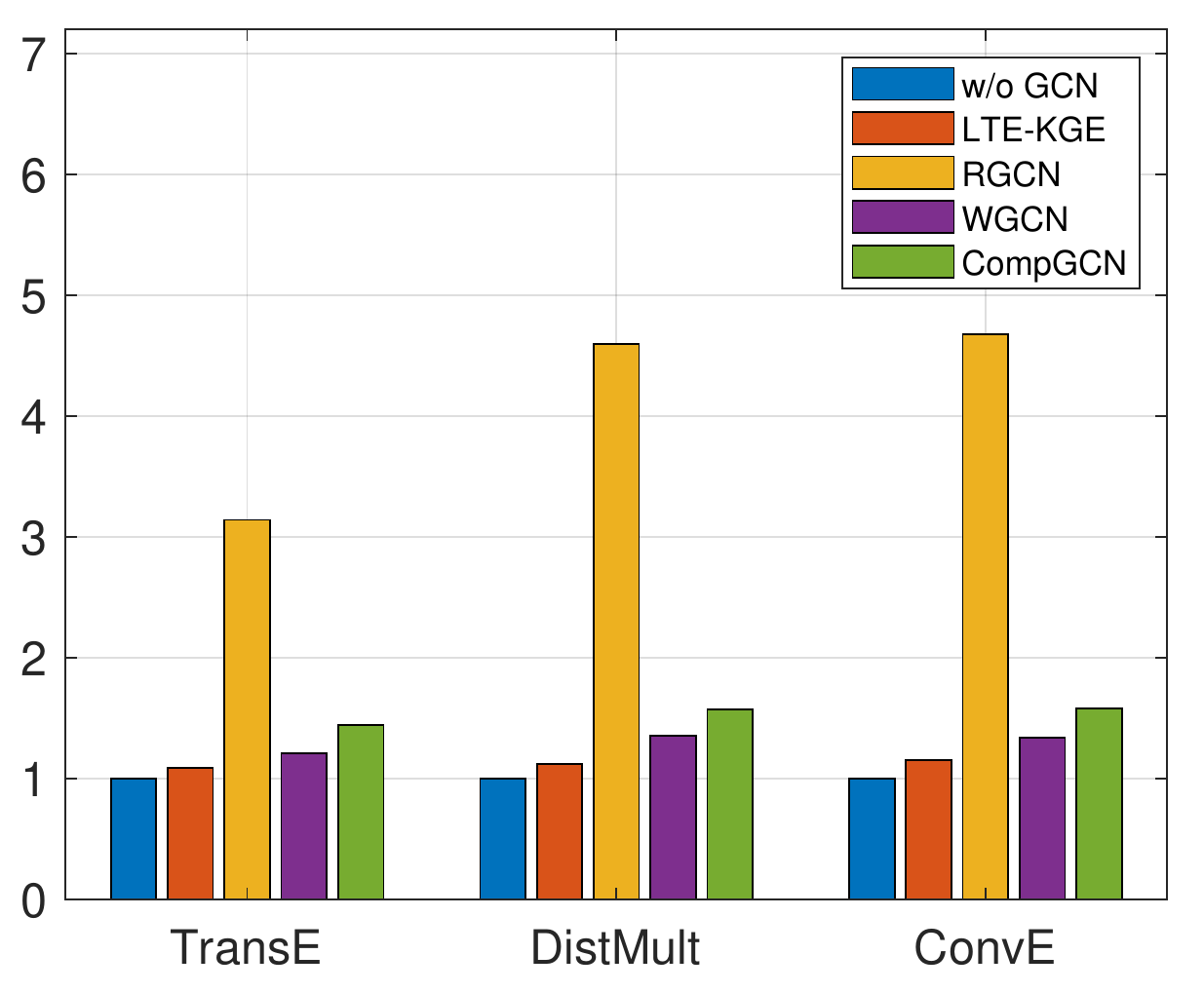}
      \caption{Test Time}\label{subfig:test_time}
    \end{subfigure}\hfil
    \vspace{-1mm}
    \caption{Comparison of training/test time. The time usage of models without GCNs is scaled to 1.}
    \label{fig:time}
    \vspace{-3mm}
\end{figure}

We also evaluate the efficiency of different models. We train and test the models on the same machine. The batch size, the number of training epochs, and the number of test samples are the same. The number of GCN layers is 1. Figure \ref{fig:time} shows the training and test time usage of different models. The results demonstrate that adding GCN encoders, especially RGCN, suffers from high training and test time usage. On the contrary, LTE-KGE models are as efficient as models without GCNs.
Therefore, LTE-KGE models have the benefit of GCN models and avoid their heavy computational load.

\subsection{Relationship between LTE-KGE and GCNs}
To understand why LTE-KGE models perform similarly to GCN-based KGC models, we show that LTE-KGE behaves like a GCN-based model with a single GCN layer being an encoder, while it does not aggregate information from neighborhoods explicitly. 

For simplicity, we only consider positive triplets and the loss function without label smoothing.
Given an entity $h$, we consider all its known neighbors $(r_i,t_i)$ such that $(h,r_i,t_i)$ are known valid triplets. Further, we assume that the formulation of LTE-KGE is $f(W\vech,\vecr,W\vect)$. Then, for a given entity $h$, our aim is to minimize the loss function
\begin{align}\label{eqn:sim_loss}
    \sum_{r\in R_h}\sum_{t:(h,r,t)\in S} \log f(W\vech,\vecr,W\vect),
\end{align}
where $R_h$ is the set consisting of all relations connected with the entity $h$, and $f$ is the scoring function of a KGE model. Note that we omit the constants involved in the formulation \eqref{eqn:sim_loss}.

When optimizing the objective, we usually use the gradient descent method to minimize the loss function \eqref{eqn:sim_loss}. The gradient with respect to $\vech$ is given as follows
\begin{align}\label{eqn:grad_h}
    \sum_{r\in R_h}\sum_{t:(h,r,t)\in S}\frac{1}{f(W\vech,\vecr,W\vect)} \frac{\partial f}{\partial \vech}(W\vech,\vecr,W\vect).
\end{align}
In our experiments, the scoring function $f$ can be TransE, DistMult, or ConvE, of which the definition are described in Section \ref{sec:pre_kgc}. In the following, we calculate the gradients for these three models.

\subsubsection{Gradients for LTE-TransE}
The gradient for LTE-TransE is
\begin{align*}
    \sum_{r\in R_h}\sum_{t:(h,r,t)\in S} \frac{1}{\|W\vech+\vecr-W\vect\|_{1/2}}\frac{\partial}{\partial \vech}\|W\vech+\vecr-W\vect\|_{1/2}.
\end{align*}
When the distance is $L_1$ norm, the above formulation is
\begin{align}\label{eqn:grad_transe_l1}
    \sum_{r\in R_h}\sum_{t:(h,r,t)\in S} W^\top \frac{I(W\vech,W\vect-\vecr)}{\|W\vech+\vecr-W\vect\|_1
     },
\end{align}
where $I(\textbf{x},\textbf{y}):\mathbb{R}^n\times \mathbb{R}^n\rightarrow\mathbb{R}^n$ is a function satisfying $[I(\textbf{x},\textbf{y})]_i$ is $-1$ if $x_i<y_i$ or otherwise $1$.
Note that the $L_1$ norm is non-differentiable at $0$. For implementation and notation convenience, we choose $\textbf{1}$ as the gradients at $0$, which belongs to the subderivative of the $L_1$ norm.

When the distance is $L_2$ norm, the gradient is
\begin{align}\label{eqn:grad_transe_l2}
    \sum_{r\in R_h}\sum_{t:(h,r,t)\in S}-\frac{W^\top}{\|W\vech+\vecr-W\vect\|_2^2
     }W\vect+\frac{W^\top(\vecr+W\vech)}{\|W\vech+\vecr-W\vect\|_2^2
     }.
\end{align}

\subsubsection{Gradients for LTE-DistMult}
The gradient  is 
\begin{align}\label{eqn:grad_distmult}
    &\sum_{r\in R_h}\sum_{t:(h,r,t)\in S}\frac{1}{((W\vech)^\top \vecR(W\vect))} \frac{\partial}{\partial \vech}((W\vech)^\top \vecR(W\vect))\nonumber\\
    =&\sum_{r\in R_h}\sum_{t:(h,r,t)\in S}\frac{W^\top\vecR}{((W\vech)^\top \vecR(W\vect))} W\vect
\end{align}

\subsubsection{Gradients for LTE-ConvE}
The gradient for LTE-ConvE is 
\begin{align*}
    \sum_{r\in R_h}\sum_{t:(h,r,t)\in S} \frac{1}{g(\vech,\vecr,\vect)}\frac{\partial}{\partial \vech}g(\vech,\vecr,\vect),
\end{align*}
where $g(\vech,\vecr,\vect)=\sigma(\text{vec}(\sigma([\bar{\vecr}, \overline{W\vech}]*\omega))W')^\top  (W\vect)$ and $W'$ is the weight matrix in ConvE. Then we have that
\begin{align*}
    \frac{\partial}{\partial \vech}g(\vech,\vecr,\vect)=\frac{\partial}{\partial \vech}\sigma(\text{vec}(\sigma([\bar{\vecr}, \overline{W\vech}]*\omega))W')^\top (W\vect),
\end{align*}
Let $\lambda(\vech,\vecr,\vect)=\frac{1}{g(\vech,\vecr,\vect)}\frac{\partial}{\partial \vech}\sigma(\text{vec}(\sigma([\bar{\vecr}, \overline{W\vech}]*\omega))W')^\top$. Then the gradient is
\begin{align}\label{eqn:grad_conve}
    \sum\nolimits_{r\in R_h}\sum\nolimits_{t:(h,r,t)\in S}\lambda(\vech,\vecr,\vect)W\vect.
\end{align}

In summary, the gradients of the $L_2$-version of LTE-TransE \eqref{eqn:grad_transe_l2}, LTE-DistMult \eqref{eqn:grad_distmult}, and LTE-ConvE \eqref{eqn:grad_conve} can be unified as
\begin{align}\label{eqn:uni_grad}
     \sum_{r\in R_h}\sum_{t:(h,r,t)\in S}a(\vech,\vecr,\vect)W\vect + b(\vech,\vecr,\vect),
\end{align}
where $a(\vech,\vecr,\vect)$ and $b(\vech,\vecr,\vect)$ are two scalar-valued functions.
We can regard the formulation \eqref{eqn:uni_grad} as an information aggregation from neighbor entities and relations. If we use gradient descent to minimize the loss function, then the optimization process is equivalent to a 1-layer GCN iteration, where the linear transformation $W$ corresponds to the weight matrices in GCN aggregation. Although the gradient of the $L_1$-version of LTE-TransE \eqref{eqn:grad_transe_l1} does not satisfy the unified formulation \eqref{eqn:uni_grad}, the optimization process also corresponds to an information aggregation based on graph structures. In summary, the combination of a KGE model and gradient descent already behaves like GCN aggregations. Thus, additional explicit aggregations are unnecessary, and it is reasonable that LTE-KGE achieves similar performance to GCN-based KGC models. 

It is worth noting that \citet{ke-gcn} also use gradient descent to discuss the relationship between KGE and GCNs. However, their main aim is to unify the GCNs by a gradient descent framework and design new aggregation schemes, while our aim is to explain the equivalence between KGE and GCN-based KGC models.

\section{Related Work}

Our work mainly focuses on GCN-based knowledge graph completion (KGC), which is relevant to the Web research community in two aspects.
First, knowledge graphs (KGs) play an important role in many applications on the Web, such as search and recommendations. Since the completeness of KGs significantly influence their applicability and manually completing KGs is expensive, how to complete KGs automatically is a fundamental research problem in the community. Second, GCNs are popular algorithms to deal with many Web-related problems, including knowledge graph completion. A better understanding of the GCNs' effects in KGC can give us inspiration and guidance to build new KGC models. Therefore, as a systematic analysis of recently proposed GCN-based KGC models, our work is relevant to the community.

Moreover, our work is related to knowledge graph embeddings, GCNs in KGC, and other work reviewing KGC models.

\udfsection{Knowledge Graph Embeddings}
Knowledge graph embedding models usually use simple architectures to measure the plausibility of triplets in knowledge graphs. For example, TransE \cite{transe}, TransH \cite{transh}, TransR \cite{transr}, and RotatE \cite{rotate} define distance-based scoring functions in either real or complex spaces. RESCAL \cite{rescal}, DistMult \cite{distmult}, ComplEx \cite{complex}, ANALOGY \cite{analogy}, TuckER \cite{tucker}, and LowFER \cite{lowfer} use tensor factorization to model the knowledge graph completion process. Their scoring functions usually consist of inner products between real or complex vectors. Some works also turn to neural networks to generate knowledge graph embeddings. For example, ConvKB \cite{convkb}, ConvE \cite{conve}, and InteractE \cite{interacte} use convolutional neural networks to 
generate more expressive entity and relation embeddings. CapsE \cite{capse} uses the capsule network to model the interaction in knowledge graphs.

\udfsection{GCNs in Knowledge Graph Completion}
Early GCNs are usually designed for homogeneous graphs \cite{gcn,graphsage}, in which nodes and edges have only one type. However, directly using these GCNs in KGC usually leads to poor performance, since knowledge graphs have different types of nodes (entities) and edges (relations). To address the problem, RGCN \cite{rgcn} extends the classical GCN and proposes to apply relation-specific transformations in GCN's aggregation. To alleviate the over-parameterization problem caused by relation-specific weight matrices, they also use basis and block-diagonal decomposition. \citet{wgcn} propose a weighted graph convolutional network, which uses learnable relation-specific scalar weights during GCN aggregation. 
However, these models do not explicitly learn relation embeddings, which are important for knowledge graph completion. To tackle this problem, many works, such as VR-GCN \cite{vec_kg}, CompGCN \cite{compgcn}, and KE-GCN \cite{ke-gcn}, propose extensions of GCNs to embed both entities and relations in knowledge graphs. Moreover, CompGCN and KE-GCN combine the power of GCNs and the strength of knowledge graph completion models to further exploit the interactions among entities and relations.

\udfsection{Reviews for KGC models}
Reviews for KGC models are constantly emerging in recent years, which help us to understand KGC models better. For example, \citet{KGE_survey} and \citet{KGE_survey2} comprehensively review the architectures and categories of existing KGC models. \citet{old_dog} conduct extensive experiments to demonstrate that early KGC models with advanced training strategies can achieve good performance. Accordingly, they suggest that many advanced architectures and techniques should be revisited to reassess their individual beneﬁts. \citet{reeval} find that some works use unfair evaluation protocols, which may lead to inflating performance. However, to our best knowledge, there exists no work that provides a comprehensive analysis for GCN-based KGC models, which is our main aim.

\section{Conclusion}
In this paper, we conduct extensive experiments to find the real critical part behind the complicated architectures of GCN-based KGC models. Surprisingly, we observe from experiments that the graph structure modeling is unimportant for GCN-based KGC models. Instead, the ability to distinguish different entities and the transformations for entity embeddings account for the performance improvements. Based on this observation, we propose LTE-KGE that enhances KGE models with linearly transformed entity embeddings, achieving comparable performance to GCN-based KGC models. Thus, we suggest that novel GCN-based KGC models should count on more ablation studies to validate their effectiveness.

\section*{Acknowledgments}
We would like to thank all the anonymous reviewers for their insightful comments. This work was supported in part by National Science Foundations of China grants 61822604, U19B2026, 61836006, and 62021001, and the Fundamental Research Funds for the Central Universities grant WK3490000004.

\bibliographystyle{ACM-Reference-Format}
\bibliography{sample-base}

\appendix

\section{Dataset Statistics}\label{app:data}
Besides FB15K-237 (FB237) and WN18RR used in the main text, we also include the results on a larger dataset YAGO3-10 \cite{yago3} in the Appendix \ref{app:rst_yago}. The statistics of these three datasets are summarized in Table \ref{table:datasets}.

\begin{table}[!ht]
\caption{Statistics of the benchmark datasets.}
\label{table:datasets}
\centering
\begin{tabular}{l *{3}{c}}
    \toprule
     & FB237 & WN18RR  & YAGO3-10 \\
    \midrule
    \#Entity  & 14,541 & 40,943 &123,182 \\
    \#Relation & 237 & 11  &37  \\
    \#Train  & 272,115 & 86,835 & 1,079,040  \\
    \#Valid  &17,535 &3,034 & 5,000 \\
    \#Test  &20,466 &3,134 &5,000\\
    \bottomrule
\end{tabular}
\end{table}

\section{Evaluation Metrics}
We use the mean reciprocal rank and Hits@N as metrics in the experiments. Their definitions are as follows.

The mean reciprocal rank is the average of the reciprocal ranks of results for a set of queries Q:
    \begin{align*}
        \text{MRR}=\frac{1}{|Q|}\sum_{i=1}^{|Q|}\frac{1}{\text{rank}_i}.
    \end{align*}
The Hits@N is the ratio of ranks that no more than $N$:
    \begin{align*}
        \text{Hits@N}=\frac{1}{|Q|}\sum_{i=1}^{|Q|}\mathds{1}_{x\le N}(\text{rank}_i),
    \end{align*}
where $\mathds{1}_{x\le N}(\text{rank}_i)=1$ if $\text{rank}_i\le N$ or otherwise $0$.

\begin{figure}[ht]
    \centering
    \begin{subfigure}{0.5\columnwidth}
      \includegraphics[width=110pt]{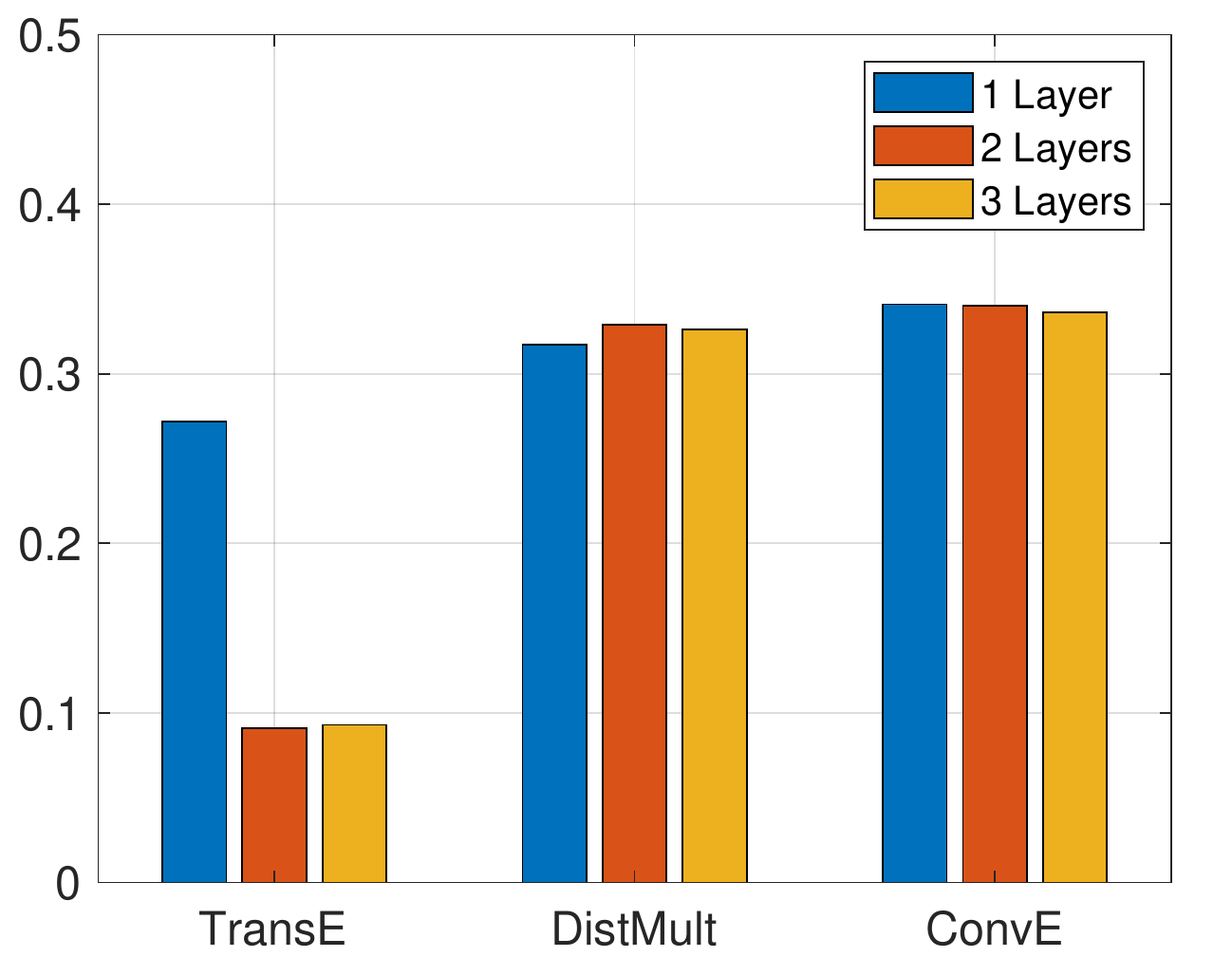}
      \caption{WGCN+X}\label{subfig:wgcn_layers}
    \end{subfigure}\hfil
    \begin{subfigure}{0.5\columnwidth}
      \includegraphics[width=110pt]{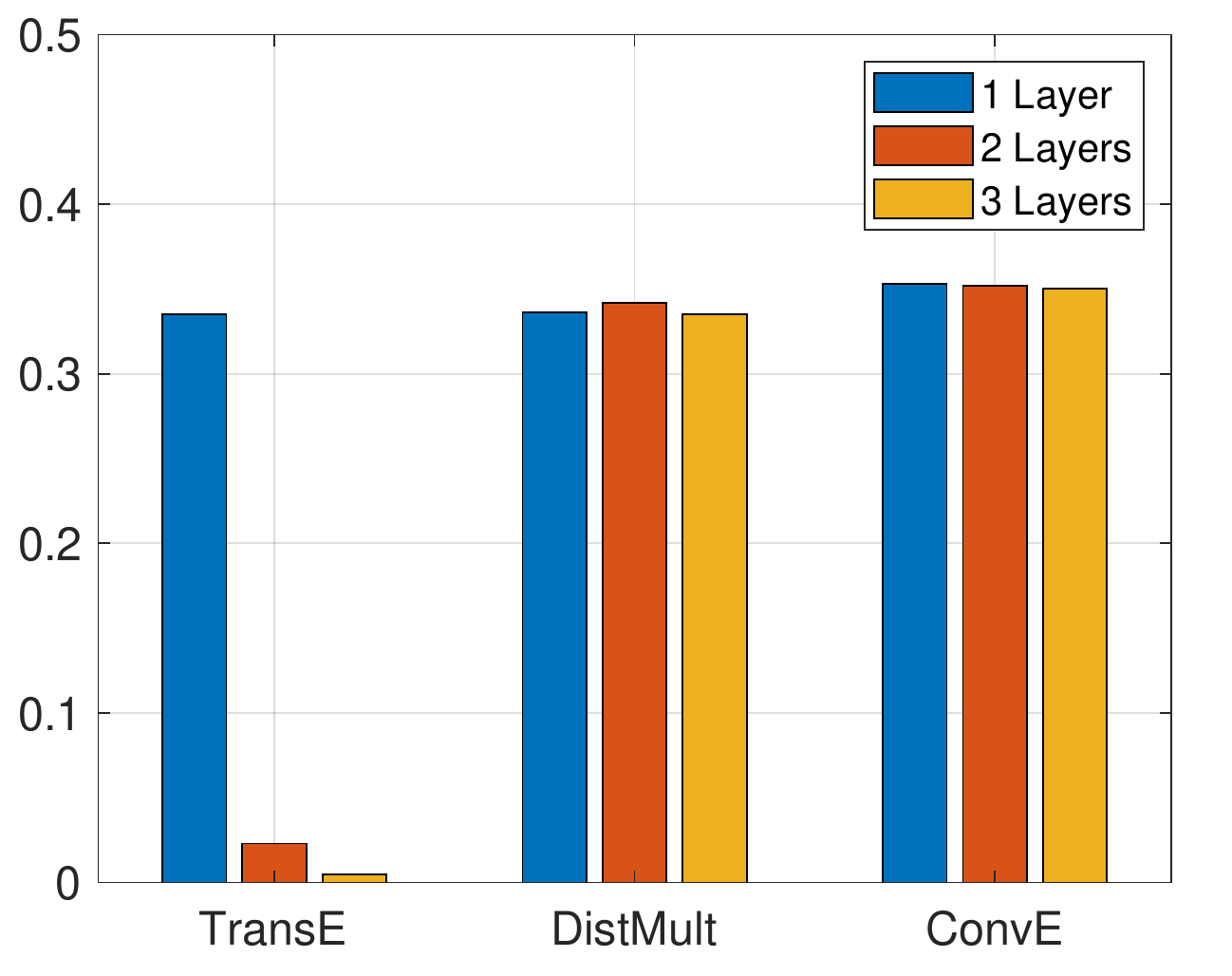}
      \caption{CompGCN+X}\label{subfig:compgcn_layers}
    \end{subfigure}\hfil
    \caption{MRR for different numbers of GCN layers on FB237 We use the models WGCN and CompGCN.}\label{fig:num_layers}
\end{figure}

\section{Number of GCN Layers} 
Many existing GCN-based knowledge graph completion models use either 1 or 2 GCN layers to achieve their best performance. To explore the effect of different numbers of GCN layers, we train WGCN and CompGCN with 1, 2, and 3 GCN layers on FB237. Figure \ref{fig:num_layers} shows that in many cases, more GCN layers do not consistently improve the performance. Sometimes the performance even decreases as the number of GCN layers increases. Since adding more GCN layers captures more global graph structures, the results that only using a single GCN layer is enough to achieve satisfying performance also suggest that graph structure modeling may be unimportant for GCN-based KGC models.

\begin{table}[h]
    \caption{\modify{MRR results on the  YAGO3-10 dataset.}}
    \label{table:yago}
    \centering
    \modify{
    \begin{tabular}{l *{1}{c}}
        \toprule
         & ConvE \\
        \midrule
        CompGCN & .348\\
        CompGCN+RAT & .337  \\
        CompGCN+WNI & .358  \\
        CompGCN+WSI &.262 \\
        CompGCN+WSI+RAT & .186\\
        \midrule
        LTE-ConvE &.362\\
        \bottomrule
    \end{tabular}
    }
\end{table}

\modify{
\section{Results on YAGO3-10}\label{app:rst_yago}
To validate our conclusions on larger datasets, we conduct experiments for the state-of-the-art CompGCN on YAGO3-10, which consists of 123k entities, 37 relations, and 1079k training triplets. The MRR results are shown in Table \ref{table:yago}. Note that: a)  we cannot find the best hyper-parameters in the paper or official implementation of CompGCN for YAGO3-10; b) tuning models on such a large dataset is expensive; c) our aim is not to provide a benchmark for existing models. Thus, we did not spend time tuning hyper-parameters, and CompGCN may not achieve its best performance. Nonetheless, we can make helpful conclusions based on the relative performance of different model variants. Table \ref{table:yago} demonstrates that randomly breaking adjacency matrices (RAT) and only using self-loop information (WNI) do not significantly affect the performance of GCN-based models.  Compared with FB15k-237, only using neighbor information (WSI/WSI+RAT) leads to lower performance than full CompGCN on YAGO3-10. It is because the average numbers of neighbor entities in YAGO3-10 and FB15k-237 are 8.8 and 18.7, respectively. It is more challenging to distinguish entities with different semantics based on fewer neighbor entities. The results are consistent with the analyses in Section \ref{sec:analysis}.
}

\end{document}